\documentclass[letterpaper, 10 pt, conference]{ieeeconf}  

\IEEEoverridecommandlockouts                              

\usepackage[pdftex]{graphicx}
\usepackage{amssymb}
\usepackage{amsmath}
\usepackage{bm}
\usepackage{cite}
\usepackage{epsfig}
\usepackage{caption}
\usepackage{color}
\usepackage[dvipsnames]{xcolor}
\usepackage{psfrag}
\usepackage{cases}
\usepackage{acronym}
\usepackage{setspace}
\usepackage{times}
\usepackage{latexsym}
\usepackage{dblfloatfix}
\usepackage[ruled,linesnumbered]{algorithm2e}
\usepackage{metalogo}
\usepackage{tabularx}
\usepackage{array}
\usepackage{multirow}
\usepackage{booktabs}
\usepackage{graphicx}
\usepackage{subfigure} 
\let\labelindent\relax
\usepackage{enumitem}

\usepackage{hyperref}
 \usepackage{threeparttable}
\hypersetup{
    colorlinks=true,
    linkcolor=blue,
    filecolor=blue,      
    urlcolor=blue,
    citecolor=cyan,
}
\usepackage{geometry}
\title{\bf SA-LOAM: Semantic-aided LiDAR SLAM with Loop Closure}

\author{Lin Li$^{1}$, Xin Kong$^{1}$, Xiangrui Zhao$^{1}$, Wanlong Li$^{2}$, Feng Wen$^{2}$, Hongbo Zhang$^{2}$ and Yong Liu$^{1,*}$
\thanks{$^{1}$Lin Li, Xin Kong, Xiangrui Zhao, and Yong Liu are with the Institute of Cyber-Systems and Control, Zhejiang University, Hangzhou 310027, P. R. China. (*Yong Liu is the corresponding author, email: yongliu@iipc.zju.edu.cn).}
\thanks{$^{2}$Wanlong Li, Feng Wen and Hongbo Zhang are with Huawei Noah’s Ark Lab, Beijing, China.}
}

\begin{document}

\maketitle
\thispagestyle{empty}
\pagestyle{empty}

\begin{abstract}

LiDAR-based SLAM system is admittedly more accurate and stable than others, while its loop closure detection is still an open issue. 
With the development of 3D semantic segmentation for point cloud, semantic information can be obtained conveniently and steadily, essential for high-level intelligence and conductive to SLAM. 
In this paper, we present a novel semantic-aided LiDAR SLAM with loop closure based on LOAM, named SA-LOAM, which leverages semantics in odometry as well as loop closure detection. Specifically, we propose a semantic-assisted ICP, including semantically matching, downsampling and plane constraint, and integrates a semantic graph-based place recognition method in our loop closure detection module. Benefitting from semantics, we can improve the localization accuracy, detect loop closures effectively, and construct a global consistent semantic map even in large-scale scenes. Extensive experiments on KITTI and Ford Campus dataset show that our system significantly improves baseline performance, has generalization ability to unseen data and achieves competitive results compared with state-of-the-art methods.

\end{abstract}

\section{Introduction}

Simultaneous Localization And Mapping (SLAM) has developed rapidly in recent decades as one of the fundamental capabilities of robotics. 
Researchers have proposed a large number of SLAM systems based on camera and LiDAR. Compared with the camera, LiDAR can obtain accurate distance measurements directly and is insusceptible from illumination changes, so LiDAR-based SLAM systems are generally more stable and precise.

LOAM\cite{zhang2014loam} is a state-of-the-art LiDAR-based SLAM system.
Based on it, many methods\cite{shan2018lego, chen2020sloam, 7139486} were proposed to improve its performance. 
However, most of them cannot eliminate the cumulative error and build a globally consistent map due to the absence of a practical loop closure detection module. 
Moreover, they operate at the low-level primitives, which is unsuitable for reliably executing high-level instructions, such as ``parking next to the pole/tree". The robots need to transform semantic concepts into a spatial representation, maintaining a high-quality semantic map beyond the map typically built by conventional SLAM. 

 \begin{figure}[t]
    \centering
    \includegraphics[width=0.9\columnwidth]{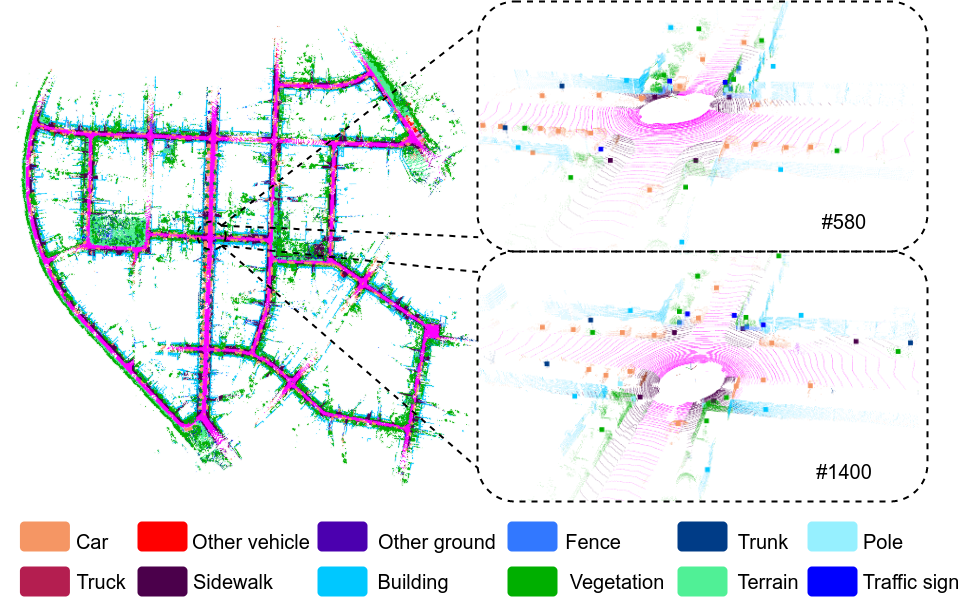}
    \caption{Semantic map built by our semantic-aided LOAM on KITTI sequence 01. Different color represents various semantics, and enlarged points are graph nodes. It shows one example of loop closure based on the semantic graph.
    }
    \label{demo}
 \end{figure}

\begin{figure*}[t]
    \centering
    \includegraphics[width=1.95\columnwidth]{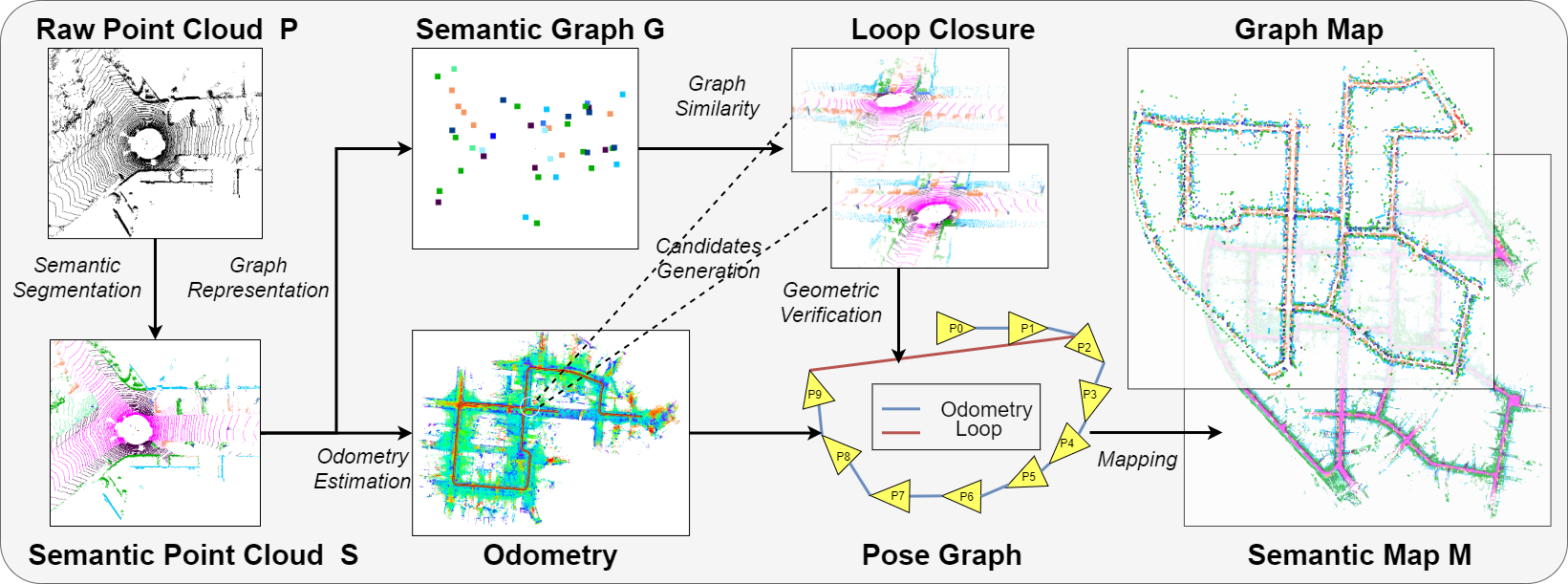}
    \caption{The pipeline of our system. Raw input point cloud $P$ is input to semantic segmentation to obtain semantic point cloud $S$. $S$ is sent to motion estimation and loop closure module simultaneously. Pose graph is updated when loop closure is detected and a global semantic map $M$ is maintained gradually. Each semantic graph $G$ together constructs a graph map for efficient loop closure detection.}
    \label{pipeline}
        \vspace{-15pt}
 \end{figure*}

With the development of deep learning, semantic information can be conveniently acquired from images, which has been used to assist SLAM like removing dynamic objects~\cite{8421015} or assisting pose estimation~\cite{10.1007/978-3-030-01225-0_15}. Unlike images, LiDAR point clouds are sparse, unstructured and lack texture features, but naturally have accurate 3D spatial representation. Thanks to the recent development of point cloud semantic segmentation~\cite{milioto2019rangenet++, hu2020randla, thomas2019KPConv, zhou2020cylinder3d, zhang12356deep, tang2020searching, cortinhal2020salsanext}, some works~\cite{chen2020sloam,chen2019suma++} are capable introduce semantics into the pure LiDAR SLAM system. 
SLOAM~\cite{chen2020sloam} is designed specially for forest inventory and SUMA++\cite{chen2019suma++} is a surfel-based method~\cite{behley2018efficient}, achieving amazing accuracy in highway scenes by dynamic removing. Moreover, semantics have shown great potential in loop closure detection~\cite{kong2020semantic, chen2020overlapnet} but not fully integrated into the LiDAR SLAM system yet. There are still few complete LiDAR-based semantic SLAM systems, thus deserve further exploration.

To tackle the above issues, we present SA-LOAM, a novel semantic-aided LOAM-based SLAM system with loop closure. Specifically, we build a semantic-assisted ICP and semantic graph-based loop closure detection module on an open-source LOAM-based SLAM system called FLOAM~\cite{wang2020floam}. 
In our semantic-assisted ICP, semantic-based clustering, downsampling, and plane constraints are implemented to reduce false matches, balance efficiency and accuracy, and remove wrong planar features to improve the registration quality. Our loop closure detection module is based on our previous work~\cite{kong2020semantic}, which turns 3D scenes into semantic graphs and obtains scenes similarity through a graph matching network. We integrate it with loop candidates generation, geometric verification, and maintain a lightweight semantic graph map for efficient and robust loop closure detection.
Fig.~\ref{demo} is a demonstration of our result. We hope to draw attention in exploiting semantic LiDAR system. Our contributions are concluded as follows:
\begin{itemize}
    \item We propose a complete LiDAR-based semantic SLAM system that can construct a global consistent semantic map even in large-scale scenes. 
        \item We present a semantic assisted ICP method based on LOAM, which makes full use of semantics and improves the accuracy of the odometry.
    \item We integrate a semantic-based loop closure detection method in our system and maintain a graph map for efficiently detecting loop closures and eliminating the accumulated error.
    \item Experiments on the KITTI and Ford Campus dataset show that our system achieves competitive performance compared with state-of-the-art methods and generally promotes the accuracy of baseline to a large margin.   
\end{itemize}

\section{Related Work}

\subsection{LiDAR SLAM}

To estimate the ego-motion, LOAM~\cite{zhang2014loam, zhang2017low} extracts edge and planar features from the raw point cloud, then minimizes the distance of point-to-line and point-to-plane. Though proposed in 2014, LOAM continually ranks the top-tier on the KITTI Odometry Benchmark\cite{geiger2013vision} among purely LiDAR-based approaches. LeGO-LOAM~\cite{shan2018lego} segments the raw point cloud, filters out small clusters, and estimates pose by a two-step Levenberg-Marquardt optimization method. Though it can filter out some dynamic objects, many stationary objects are removed unwillingly, resulting in useful feature loss. IMLS-SLAM~\cite{deschaud2018imls} uses a scan-to-model framework, leveraging Implicit Moving Least Squares (IMLS) to reconstruct the plane, and minimizes the distance of points to the IMLS surface. However, it cannot achieve real-time performance due to the slow process of calculating IMLS.

Recently, a few researchers combine semantic information to assist point cloud registration. Zaganidis et al.~\cite{8387438} use semantic information to extend Normal Distributions Transform (NDT)~\cite{biber2003normal} and Generalized Iterative Closest Point (GICP)~\cite{segal2009generalized}. SLOAM~\cite{chen2020sloam} uses semantic segmentation to detect trunks and ground, then models them separately, achieving good results in forest scenes. SUMA~\cite{behley2018efficient} is a surfel-based approach with loop closure, and SUMA++~\cite{chen2019suma++} extends it with semantics to filter out dynamic objects, achieving state-of-the-art performance.

\subsection{LiDAR-based Loop Closure}

Existing LiDAR-based loop closure detection methods can be divided into local, global, and segment-based methods. 

Local descriptor methods  ~\cite{johnson1999using, salti2014shot, tombari2011combined, guo2019local} usually define a local reference frame for each feature point, and then they use point distribution information to describe features. FPFH~\cite{rusu2009fast} uses normal to construct histograms for key points, effectively describing the local geometry around each key points. However, such local features are not discriminative enough to distinguish similar local structures.

Global descriptors are produced from the whole LiDAR scan. M2DP~\cite{he2016m2dp} projects 3D points to a series of 2D planes, generates signature matrixes and regards the singular vectors as global features.
Scan Context~\cite{Kim2018ScanCE}, LocNet~\cite{8500682}, and ISC~\cite{9196764} convert the 3D point cloud to a 2D image representation, then compare those 2D images to find loops. PointNetVLAD~\cite{angelina2018pointnetvlad}, SeqLPD~\cite{liu2019seqlpd} and LPD-Net~\cite{liu2019lpd} extract point features with PointNet~\cite{qi2017pointnet, qi2017pointnet++} and aggregate global features via NetVLAD~\cite{arandjelovic2016netvlad}. 
OverlapNet~\cite{chen2020overlapnet} projects the 3D points to a range image including multiple clues (e.g., range, normal, intensity, and semantic), then a siamese network is used to calculate the overlap score between input scans. These methods are typically sensitive to view-point changes and partial loss of points.

Recently, some segmentation-based approaches have been proposed. SegMatch~\cite{dube2017segmatch} segments point clouds into distinct elements, extracts their features, and matches corresponding elements by a random forest. SegMap~\cite{segmap2018, dube2019segmap} uses 3D CNNs to learn descriptors for 3D point cloud segments. Benefiting from the robustness of semantics, our prior work~\cite{kong2020semantic} semantically segments the point clouds and constructs them as semantic graphs, regarding semantic objects as graph nodes. A graph similarity network is adopted to capture their relationships and obtain the similarity scores for graph pairs. More similar to human perception, operating at semantic or objects level is more robust to occlusion and view-point changes, especially in reverse loop closure detection.

\section{Methodology}

\subsection{System Overview}
In this part, we give an overall introduction to the proposed semantic-aided LiDAR SLAM system. Different from the existing methods~\cite{chen2019suma++, chen2020sloam}, which focus on fusing semantics in registration, we explore more usage of semantics in our system, mainly including semantic-assisted ICP~\ref{ssec:icp} and semantic graph-based loop closure detection~\ref{ssec:loop}. 

Fig.~\ref{pipeline} is the pipeline of our approach. The raw point cloud $P$ is first sent into the off-the-shelf semantic segmentation methods~\cite{milioto2019rangenet++, hu2020randla, thomas2019KPConv, zhou2020cylinder3d, zhang12356deep, tang2020searching, cortinhal2020salsanext} to obtain semantic point cloud $S$ with point-wise class label \(l \in L\), and \(L\) is the number of semantic categories. Then $S$ is sent to the odometry estimation module to extract flat planar features and sharp edge features, which are later used to estimate the odometry via registered with a local map.  At the same time, the semantic point cloud $S$ is converted to semantic graph representation $G$, which is further input to the loop closure detection module. When detecting a closed loop, we will update the pose graph and optimize poses. Thus, we can incrementally obtain a high-quality global semantic map $M$.

\subsection{Semantic Assisted ICP}\label{ssec:icp}
\begin{figure}[!t]
    \centering
    \subfigure[FLOAM]{
        \centering
        \includegraphics[width=0.40\columnwidth]{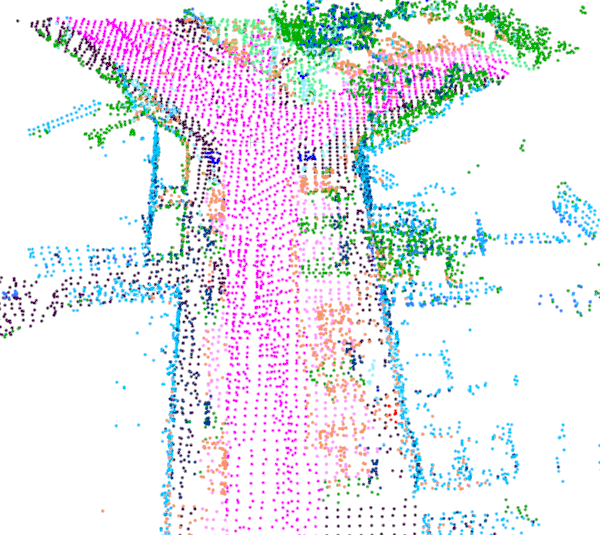}
            }
    \subfigure[Ours]{
        \centering
        \includegraphics[width=0.40\columnwidth]{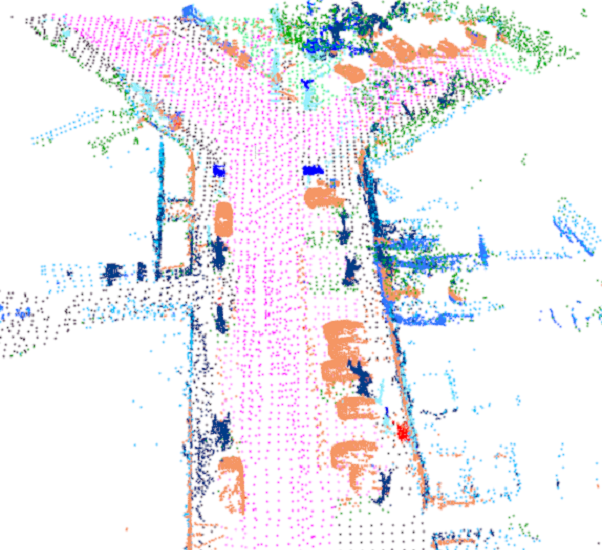}
        }
    \caption{Visualization of proposed semantic-based downsampling. Most methods use voxel-grid downsampling, causing information loss for small objects while ours effectively retain valuable small object points.}
    \label{pic:filter}
 \end{figure}

LOAM uses edge and planar features to register point clouds, achieving accurate and fast localization and mapping. We extend this method with semantic information, and the specific improvements are as follows:

\begin{itemize}
    \item Firstly, inspired by SE-NDT\cite{8387438}, we divide the features by their semantic labels and registration separately, reducing the probability of false matches. 
    \item Secondly, we downsample the point clouds separately according to their semantic labels. Most of the existing methods downsample the point clouds through voxel grid filtering for efficiency. However, some small objects containing useful information will inevitably be filtered out. Our semantic-based downsampling uses different sampling rates for different kinds of objects and can effectively maintain the information of small objects, depicted in Fig.~\ref{pic:filter}.
    \item Thirdly, we leverage semantics to constrain the plane fitting. Based on the assumptions that the ground plane should be parallel to the horizontal surface and be perpendicular to the building surface, we can remove the poorly fitted plane. 
\end{itemize}


    


\smallskip\noindent\textbf{Feature Extraction.} We adopt the feature extraction strategies of LOAM\cite{zhang2014loam}. Specifically, we extract edge and planar features from the point cloud and calculate the roughness of each point by
\begin{equation}
    roughness=\frac{1}{n\cdot\Arrowvert p_i \Arrowvert}\Arrowvert \sum_{j=1,j\neq{i}}^n(p_j-p_i) \Arrowvert
\end{equation}
where \(p_i\) represents the target point, and \(p_j\) 
is in the same ring. Then, a threshold \(\alpha\) is used to divide those points into edge points (roughness larger than \(\alpha\)) and plane points (roughness smaller than \(\alpha\)). For uniform sampling, each ring is divided into \(N_r\) parts. We select \(N_e\) sharpest points as edge features and \(N_p\) flattest points as planar features in each part.
Finally, we get edge feature points \(C_e\) and planar feature points \(C_p\) of the whole target frame.

\smallskip\noindent\textbf{Motion Estimation.} To estimate the ego-motion of LiDAR, we minimize the distance from the target edge feature point to its corresponding line and the distance from the target planar feature point to its corresponding plane. As different semantic classes have diverse meanings which are supposed to contribute to the localization discriminatively, we attach a semantic-related weight $w^l$ to each error term. Our total optimization goal is:
\begin{equation}
    r=\sum^{L}_{l}\{\sum_{i}(w^ld^{l}_{ei})+\sum_{j}(w^ld^{l}_{pj})\}
    \label{equ:opt}
\end{equation}
where \(d^{l}_{ei}\) is the distance from the $i$-th edge feature point \(c_{ei}^l\in{C_e}\) to its corresponding line, \(d^{l}_{pj}\) is the distance from the $j$-th planar feature point \(c_{pj}^l\in{C_p}\) to its corresponding plane, \(l\in L\) is the semantic label of the target feature point, \(w^l\) is a semantically related weight. 

We firstly use the most recent \(N_m\) frames of edge feature points \(C_e\) and planar feature points \(C_p\) to establish an edge feature submap \(M_e\) and a planar feature submap \(M_p\). 
In order to speed up the registration, we downsample the submap $M_e, M_p,$ and the target point cloud $C_e, C_p,$ according to their semantic labels. Fig.~\ref{pic:filter} depicts the visualization of our downsampling strategy. Then the odometry is estimated by registering the downsampled feature point cloud with the submap. Specifically, given an edge feature point \(c_{ei}^l\in{C_e}\) with it's semantic label \(l \in L\), we choose a semantic edge submap \{\(M^{l}_{e} \mid M^{l}_{e} \subset M_{e}, l \in L \)\} which has the same semantic label \(l\). We use a kd-tree to find the \(N_{ne}\) points closest to the target point as the corresponding points. These \(N_{ne}\) points are used to fit a line, and the distance from the target point to the corresponding line is used as the optimization target. Supposing \(m_{e1}^l, m_{e2}^l \in M^{l}_{e}\) are two different points on the line, the distance from the target point to the line can be calculated as follows:
\begin{equation}
    d_{ei}^l=\frac{\Arrowvert (c_{ei}^l-m_{e1}^l)\times(c_{ei}^l-m_{e2}^l)\Arrowvert }{\Arrowvert m_{e1}^l-m_{e2}^l \Arrowvert}
\end{equation}

Similarly, given a planar feature point \(c_{pi}^l \in C_p\), we find \(N_{np}\) closest points in its corresponding semantic submap \(M^{l}_{p}\). Then we use 
these \(N_{np}\) points to fit a plane.
We will reserve ground planes if their normal is vertical and building planes if their normal is horizontal. Through this constraint, the poorly fitted planes can be removed. Supposing \(m_{p1}^l, m_{p2}^l, m_{p3}^l \in M^{l}_{p}\) are three different points on the plane, the distance from \(c_{pi}^l\) 
to the plane is:
\begin{equation}
    d_{pi}^l=\frac{\Arrowvert (c_{pi}^l-m_{p1}^l)^T((m_{p1}^l-m_{p2}^l)\times(m_{p1}^l-m_{p3}^l)) \Arrowvert}{\Arrowvert (m_{p1}^l-m_{p2}^l)\times(m_{p1}^l-m_{p3}^l) \Arrowvert}
\end{equation}

\subsection{Semantic Based Loop Closure Detection}\label{ssec:loop}
Our loop closure detection module mainly consists of loop candidate generation, similarity scoring, geometric verification, and pose graph optimization. The candidate generation part proposes potential loop candidates according to odometry. Similarity scoring is based on our prior work~\cite{kong2020semantic}, which quickly estimates the similarities of pair semantic graphs through a graph similarity network, robust to occlusion and viewpoint changes. Then, geometry verification uses ICP to exclude the possible wrong proposals, avoid catastrophic false-positive loop closures. Finally, more accurate poses are updated through pose graph optimization.

\smallskip\noindent\textbf{Loop Candidates Generation.} Given the current frame point cloud, comparing it with all other point cloud frames is extremely time-consuming. Therefore, frames closer to the current position are preferred according to odometry. Specifically, frames distance less than \(\delta_{d}\) m from the current frame are kept, which is defined as:
\begin{equation}
    \delta_{d} = max(\delta_{min}, min(\delta_{max}, \sigma d))
    \label{equ:distance}
\end{equation}
where \(\sigma\) is a constant representing the drift rate of the odometry, and $d$ represents the travel distance between the candidate frame and current frame. \(\delta_{max}\) and \(\delta_{min}\) are the upper and lower bounds of $d$, as $d$ is continually increasing through the movement. We select a maximum of \(N_{candi}\) candidates. If the number of candidates filtered by Eq.~\ref{equ:distance} is greater than \(N_{candi}\), we will randomly 
select \(N_{candi}\) candidates from them. Random selection is more reasonable than selecting the frames closest to the current frame due to the existence of the cumulative error. Surely, more sampling strategies can also be tried.

\begin{table*}[t]\footnotesize
    \caption{Parameters in Our System}
    \label{table:param}
    \begin{center}
    \begin{tabular}{ccccccccccccccccc}
    \hline
    Parameter & \(\alpha\) & \(N_r\) & \(N_e\) & \(N_p\) & \(N_m\) & \(N_{ne}\) & \(N_{np}\) & \(\sigma\) & 
    \(\delta_{min}\)& \(\delta_{max}\)&\(N_{candi}\)&\(\zeta\)&\(N_{loop}\)&\(\delta_{r}\)\\
    \hline
    Value& 0.1& 6& 20& 50& 20 & 5& 5& 0.05& 15& 100 & 64& 0.95& 5& 100 \\
    \hline
    \end{tabular}
    \end{center}
        \vspace{-15pt}
\end{table*}

        
\smallskip\noindent\textbf{Similarity Scoring.} We adopt our previous semantic graph-based place recognition 
method~\cite{kong2020semantic} to calculate the similarity scores among scene pairs, which runs quickly and is lightweight and robust. We firstly cluster the semantic point cloud $S$ to semantic graph representation $G$, which is saved with its corresponding pose in the global semantic graph map for reuse. Note that each semantic graph only occupies 100x4 floats at most (maximum 100 nodes per frame with center coordinate x, y, z, and semantic label)~\cite{kong2020semantic}, which can be stored and read efficiently. According to the candidates proposed by loop candidates generation, we handily index all corresponding semantic graphs in the semantic graph map and create a batch of semantic graph pairs. Those graph pairs are processed in parallel by the graph matching network~\cite{kong2020semantic} to obtain similarity scores. We keep pairs with scores greater than \(\zeta\) and choose the \(N_{loop}\) with the highest scores from the retained pairs as the final candidates.

\smallskip\noindent\textbf{Geometric Verification.} Usually, the candidate loop picked out based on similarity is not necessarily correct. Therefore, we need to check the consistency between the current and the candidate point cloud. Similar to the odometry module~\ref{ssec:icp}, we use continuous scans near the candidate point cloud to construct a submap and use the semantic assisted ICP to register the current point cloud with the submap. If the ICP loss $r$ in Eq.~\ref{equ:opt} is larger than \(\delta_{r}\), we consider this candidate a false positive and discard it.

\smallskip\noindent\textbf{Pose Graph Optimization.} When we find a loop closure, we build a pose graph and optimize it with g2o\cite{inproceedings}. Assume that the poses corresponding to each frame of point clouds are \(\{T_0,T_1,\cdots,T_n |T_i\in SE3\}\), the error function between frame \(i\) and frame \(j\) is:
\begin{equation}
    e_{i,j}=Z^{-1}_{ij}(T^{-1}_i\cdot T_j)
\end{equation}
where \(Z_{ij}\) represents the measured relative pose between frame \(i\) and frame \(j\), obtained from the Geometric Verification module. After optimization, we get a more accurate pose for each frame, and they contribute to constructing a globally consistent semantic map $M$.


\begin{table*}[htb]

    \caption{Relative Pose Error on KITTI Odometry}
    \label{table:kitti_result_relative}
    \begin{center}
    \begin{threeparttable}
    \begin{tabular}{c c c c c c c c c}
    \hline
    Sequence & LOAM* & FLOAM & ISC-LOAM & SUMA & SUMA++ & Ours-ODOM & Ours-LOOP\\
    \hline
    00* & -/0.78 & 0.43/0.92 & 0.42/1.02 & 0.32/0.77 & 0.22/0.65 & 0.25/\textbf{0.59} & \textbf{0.18}/\textbf{0.59}\\
    01 & -/\textbf{1.43} & 0.60/2.80 & 0.63/2.92 & 0.76/11.15 & \textbf{0.47}/1.63 & 0.48/1.89 & 0.48/1.89\\
    02* & -/0.92 & 0.52/1.56 & 0.54/1.67 & 0.93/2.93 & \textbf{0.14}/3.54 & 0.28/\textbf{0.77} & 0.27/0.79\\
    03 & -/0.86 & 0.66/1.09 & 0.72/1.15 & 0.61/1.25 & 0.47/\textbf{0.67} & \textbf{0.46}/0.87 & \textbf{0.46}/0.87\\
    04 & -/0.71 & 0.52/1.43 & 0.56/1.50 & \textbf{0.27}/0.86 & \textbf{0.27}/\textbf{0.34} & 0.35/0.59 & 0.35/0.59\\
    05* & -/0.57 & 0.36/0.79 & 0.37/0.81 & 0.32/0.56 & 0.19/0.40 & 0.24/0.45 & \textbf{0.16}/\textbf{0.37}\\
    06* & -/0.65 & 0.39/0.72& 0.41/0.76 & 0.51/0.64 & 0.27/\textbf{0.47} & 0.25/0.52 & \textbf{0.24}/0.52\\
    07* & -/0.63 & 0.39/0.54 & 0.43/0.56 & 0.37/0.47 & 0.28/\textbf{0.39} & 0.22/0.41 & \textbf{0.18}/0.41\\
    08* & -/1.12 & 0.46/1.11 & 0.50/1.20 & 0.40/1.06 & 0.34/1.01 & 0.27/0.85 & \textbf{0.25}/\textbf{0.84}\\
    09* & -/0.77 & 0.55/1.28 & 0.59/1.40 & 0.41/0.79 & \textbf{0.20}/\textbf{0.58} & 0.28/0.68 & 0.25/0.77\\
    10 & -/0.79 & 0.58/1.77 & 0.62/1.87 & 0.44/0.99 & \textbf{0.30}/\textbf{0.67} & 0.35/0.78 & 0.35/0.78\\
    \hline
    Average & -/0.84 & 0.49/1.27 & 0.52/1.35 & 0.48/1.95 & \textbf{0.28}/0.94 & 0.31/\textbf{0.76} & \textbf{0.28}/\textbf{0.76}\\
    \hline
    \end{tabular}
    \begin{tablenotes} 
        \footnotesize
		\item Mean relative pose error over trajectories of 100 to 800 m: relative rotational error in degrees per 100 m / relative translational error in \%.
		Sequences marked with * contain loop closures. Bold numbers indicate the best performance in terms of translational error and rotation error. The results of LOAM marked with * is from the original paper~\cite{zhang2017low}.  
     \end{tablenotes}
     \end{threeparttable}
    \end{center}
        \vspace{-15pt}

    \end{table*}


 \begin{figure}[t]
    \centering
        \centering
        \includegraphics[width=0.7\columnwidth]{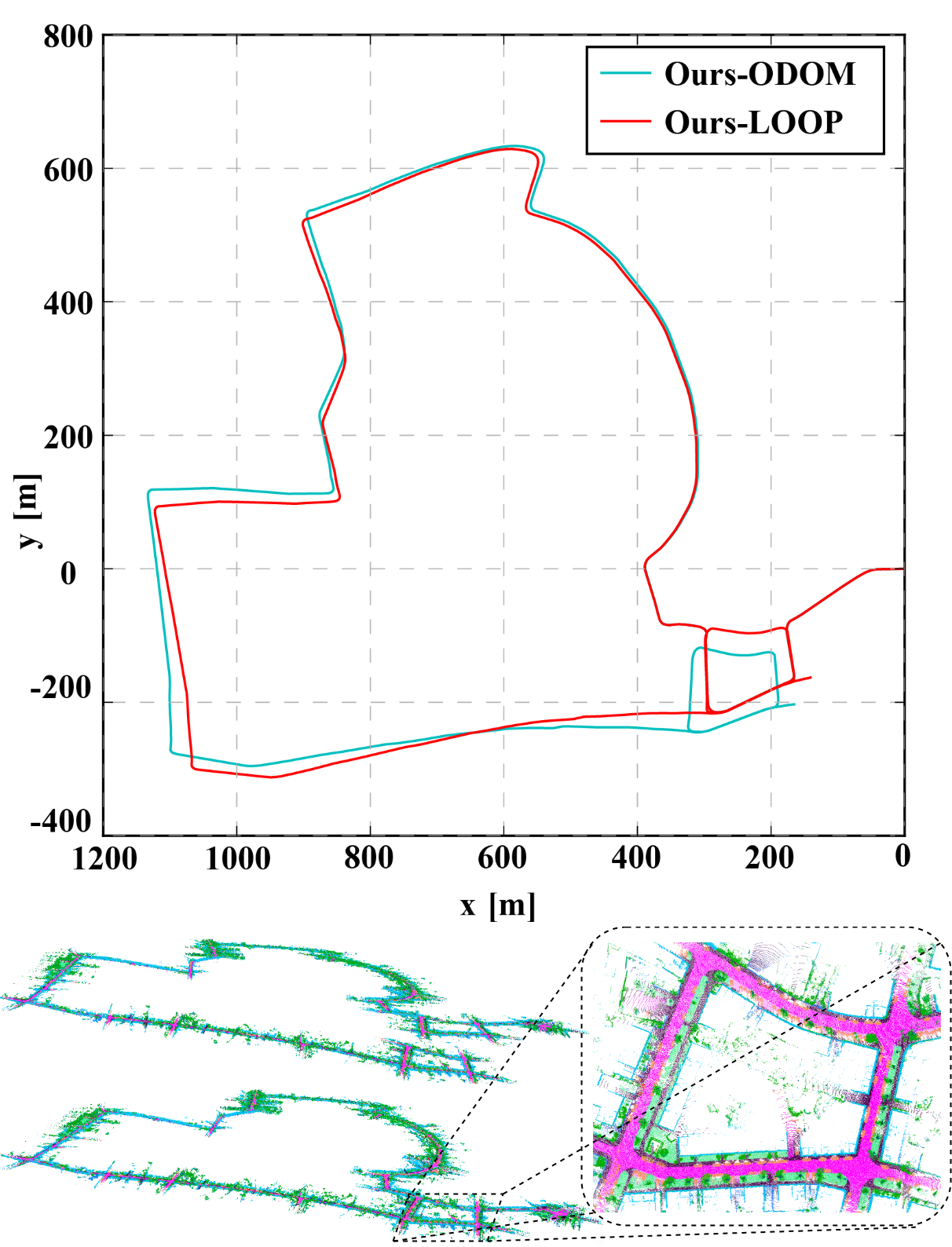}
    \caption{Trajectories and semantic maps on test sequence 19 of the KITTI dataset.}
    \label{pic:map_19}
 \end{figure}

\section{Experiments}

In this section, we design experiments to prove that (1) our semantic assisted ICP can effectively improve the odometry accuracy, (2) our semantic-based loop closure detection module can efficiently reduce the cumulative error and help establish a globally consistent map, (3) and our system has generalization ability to unseen data.

\subsection{Dataset and Implementation Details}

We evaluate our approach on KITTI Odometry Benchmark~\cite{geiger2013vision} and Ford Campus Vision and Lidar Dataset~\cite{pandey2011ford}. 
The KITTI Odometry Benchmark consists of 11 sequences (00-10) with ground truth trajectories, and the Ford Campus dataset contains two sequences (01-02). Though both datasets were collected with a Velodyne HDL-64E, the data distribution of the two data sets is quite different, as KITTI 
was collected in Germany and Ford was collected in the United States.

We choose the open-source work FLOAM\cite{wang2020floam} as our system baseline. In the semantic segmentation, we use the pre-trained model\footnote{\url{https://github.com/PRBonn/lidar-bonnetal/tree/master/train/tasks/semantic}} of RangeNet++\cite{milioto2019rangenet++} to perform semantic segmentation in line with SUMA++\cite{chen2019suma++} for a fair comparison. Both RangeNet++ and the semantic-graph-based loop closure detection method~\cite{kong2020semantic} we adopted are implemented on PyTorch~\cite{paszke2019pytorch}. RangeNet++ is trained on SemanticKITTI~\cite{behley2019semantickitti}, outputting 19 semantic classes, and we reserve 12 classes according to~\cite{kong2020semantic} as shown in Fig.~\ref{demo}, getting rid of useless semantics like the person, pedestrian. We test our method on an Intel Core i7-9750H @3.00GHz with 16 GB RAM and an Nvidia GeForce GTX1080 GPU with 8 GB RAM. The parameters we set are listed in Tab.~\ref{table:param}. Semantic-related weight $w^l$ is defined equally in our experiment, and we wish to find a reasonable automatic way for weights setting in the future. Note that we have no intention of exhaustive parameter choosing, and we believe fine-tuning will bring certain improvements. 

To evaluate how each of the proposed components contributes to improving the performance, we present poses based on semantic assisted odometry~\ref{ssec:icp} (Ours-ODOM) and poses after further optimization by loop closure~\ref{ssec:loop} (Ours-LOOP), respectively. Besides, We compare our approach with several state-of-the-art pure LiDAR-based SLAM methods such as LOAM~\cite{zhang2014loam, zhang2017low}, FLOAM~\cite{wang2020floam} (our baseline), SUMA~\cite{behley2018efficient}, SUMA++~\cite{chen2019suma++} and ISC-LOAM~\cite{9196764}. As the code of LOAM~\cite{zhang2017low} is unavailable, we directly quote its results from paper, while the results of FLOAM\footnote{\url{https://github.com/wh200720041/floam}}, ISC-LOAM\footnote{\url{https://github.com/wh200720041/iscloam}}, SUMA\footnote{\url{https://github.com/jbehley/SuMa}}, and SUMA++\footnote{\url{https://github.com/PRBonn/semantic_suma}} are obtained by using their open-source codes.

    \begin{table}[t]\footnotesize
        \caption{ \centering Absolute Trajectory Error on KITTI Odometry with Loop-closure}
        \label{table:kitti_result_absolute}
        \begin{center}
        \begin{threeparttable}
            {
        \begin{tabular}{c c c c| c c c}
        \hline
        \multirow{3}{*}{\scriptsize Sequence} & {\scriptsize ISC-} & \multirow{2}{*}{\scriptsize SUMA} & \multirow{2}{*}{\scriptsize SUMA++} & {\scriptsize Ours-} & {\scriptsize Ours-} & \multirow{3}{*}{\scriptsize Improve}\\
         & {\scriptsize LOAM} &   &   & {\scriptsize ODOM} & {\scriptsize LOOP} & \\
         & {\scriptsize (m)} &  {\scriptsize (m)}  &   {\scriptsize (m)} &  {\scriptsize (m)} &  {\scriptsize (m)} & \\
        \hline
        00* & 1.60 & 1.14 & 1.17 & 5.14 & \textbf{0.99} & 80.7\%\\
        02* & \textbf{4.77} & 44.13 & 12.99 & 9.71 & 9.24 & 4.8\%\\
        05* & 2.49 & 0.86 & \textbf{0.64} & 3.04 & 0.75 & 75.3\%\\
        06* & 1.03 & 0.68 & \textbf{0.56} & 0.69 & 0.64 & 7.2\%\\
        07* & 0.56 & 0.40 & 0.37 & 0.53 & \textbf{0.36} & 32.0\%\\
        08* & 4.88 & \textbf{2.09} & 2.44 & 3.61 & 3.24 & 10.2\%\\
        09* & 2.31 & 3.88 & \textbf{1.19} & 1.74 & 1.20 & 31.0\%\\
        \hline
        {\scriptsize Average} & 2.52 & 7.59 & 2.76 & 3.49 & \textbf{2.34} & 32.9\%\\
        \hline
        \end{tabular}
        }
        \begin{tablenotes} 
            \footnotesize
            \item  The last column represents the improvements of Our-LOOP compared to Ours-ODOM.
         \end{tablenotes}
        \end{threeparttable}
        \end{center}
        \end{table}
        
\subsection{KITTI Odometry Benchmark}

\smallskip\noindent\textbf{Relative Pose Error.} To evaluate the accuracy improvement of odometry, we test on KITTI Odometry Benchmark and use mean Relative Pose Error (mRPE) as the metric.

Tab.~\ref{table:kitti_result_relative} shows the experimental results on KITTI Odometry Benchmark. Compared with the baseline method FLOAM, our method greatly improves the odometry accuracy. LOAM has the highest accuracy among the existing LOAM-based methods, and our method exceeds LOAM on most sequences. SUMA and SUMA++ are surfel-based, and SUMA++ is the state-of-the-art semantic-based method. Compared with SUMA++, our method has better performance on long-distance sequences such as 00, 02, 05, 08. Note that sequences 00, 02, 05, 06, 07, 08, 09 have loops, and Ours-LOOP performs generally better than Ours-ODOM without surprise. Moreover, we find that loop closure optimization can significantly reduce the relative rotational error on most sequences, while has less impact on the relative translational error, which means rotational error have a greater impact on the inconsistency of the trajectory.


\smallskip\noindent\textbf{Absolute Trajectory Error.} To evaluate the ability of reducing the cumulative error and establishing a globally consistent map, we further evaluate the Absolute Trajectory Error (ATE) and give qualitative results. 

In Tab.~\ref{table:kitti_result_absolute}, we find that Ours-LOOP has a great improvement in ATE compared to Ours-ODOM on every sequence, especially on sequences 00 and 05, where Ours-LOOP reduces the ATE by 80.7\% and 75.3\% , respectively. Compared with state-of-the-art LiDAR-based methods with loop closure detection that are ISC-LOAM, SUMA, SUMA++, Ours-LOOP has a smaller ATE on most sequences and outperforms other methods in average. 

Fig.~\ref{pic:map_19} shows trajectories and semantic maps on test sequence 19 of the KITTI dataset before and after loop closure. There are global closure loops between the start and end of this trajectory. Due to the accumulated error of the odometry, the trajectory generated by Ours-ODOM is inconsistent near the start and end points. By contrast, Ours-LOOP can detect those loops and eliminate the accumulated error of the odometry to produce a globally consistent map that is depicted clearly in Fig.~\ref{pic:map_19}.




    \begin{table}[t]\footnotesize
        \caption{Absolute Trajectory Error on Ford Campus}
                \vspace{-5pt}
        \label{table:ford_abs}
        \begin{center}
        \begin{tabular}{c| c c| c}
        \hline
        \multirow{2}{*}{Algorighm} & Seq01 & Seq02 & Average \\
         & (m) & (m)  & (m)  \\
        \hline
        FLOAM & 1.61 & 7.63 & 4.62 \\
        ISC-LOAM & 2.30 & 10.38 & 6.34\\
        SUMA & 4.45 & 75.65 & 40.05\\
        SUMA++ & 4.22 & 63.27 & 33.75\\
        Ours-ODOM & \bf{1.35} & 9.02 & 5.18\\
        Ours-LOOP & \bf{1.35} & \bf{6.97} & \bf{4.16}\\
        \hline
        \end{tabular}
        \end{center}
        \vspace{-5pt}
        \end{table}
        
\subsection{Ford Campus Vision and Lidar Dataset}
In order to verify the generalization ability to unseen data, we conduct experiments on the Ford Campus dataset~\cite{pandey2011ford} and evaluate RTE as well as ATE. All models used in our method are pre-trained on KITTI and have never seen Ford's data during the training phase.

\begin{figure}[t]
    \centering
        \includegraphics[width=0.9\columnwidth]{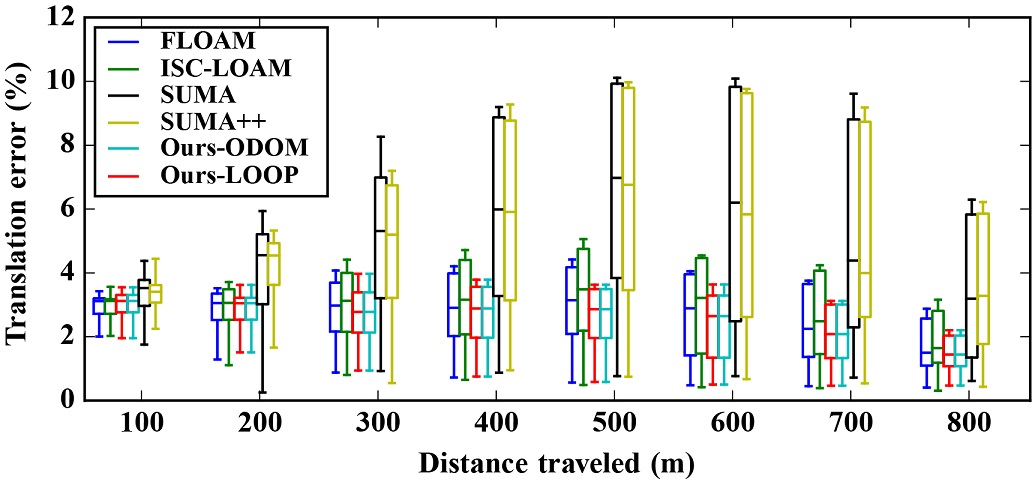}
        \vspace{-5pt}
    \caption{Relative translational error on sequence 01 of Ford dataset. Our method has a smaller relative error than others.}
    \label{pic:ford_01}
 \end{figure}
 
Fig.~\ref{pic:ford_01} shows the relative translational error on sequence 01 of the Ford dataset. As all methods' parameters are tuned on KITTI, they generally perform worse on Ford. However, the SUMA-based methods are more severely affected as we find the projection images are sparser than on KITTI due to the different sensor settings. Among all the methods, Ours-LOOP achieves the smallest relative error. 

\begin{figure}[!t]
    \centering
        \includegraphics[width=0.8\columnwidth]{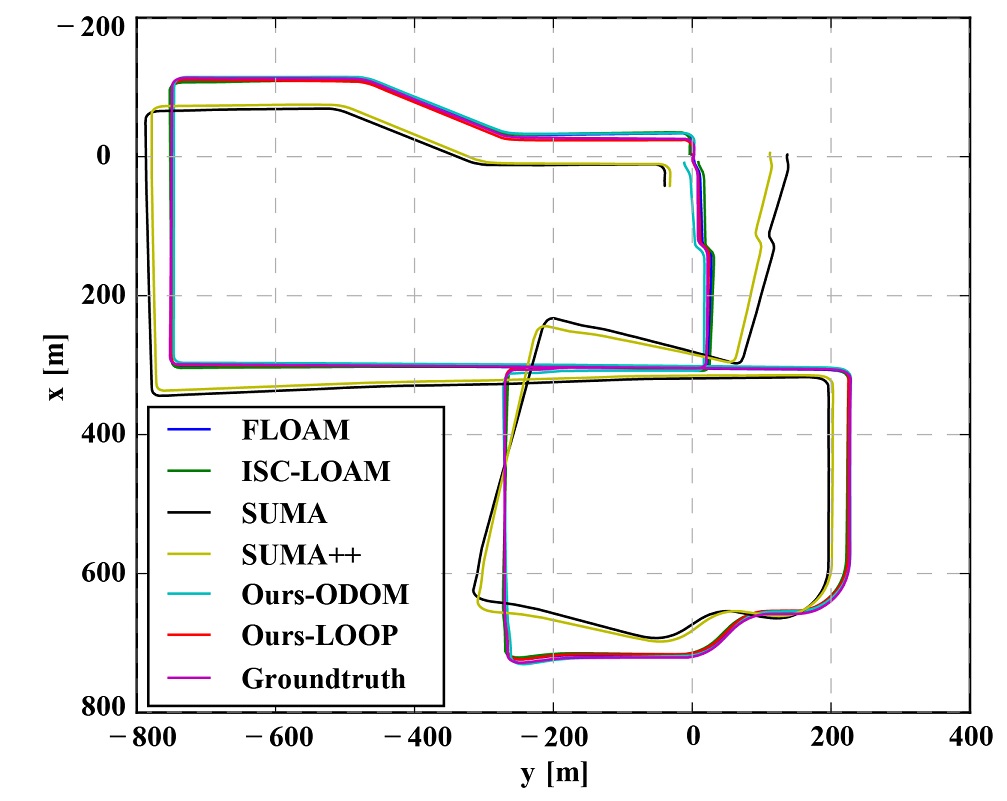}
                \vspace{-5pt}

    \caption{Trajectories on sequsence 02 of Ford dataset. Only Ours-LOOP correctly finds closed loops.}
    \label{pic:ford_relative}
        \vspace{-5pt}
\end{figure}
 
\begin{figure}[!t]
    \centering
    \subfigure[KITTI sequence 13]{
        \centering
        \includegraphics[width=0.45\columnwidth]{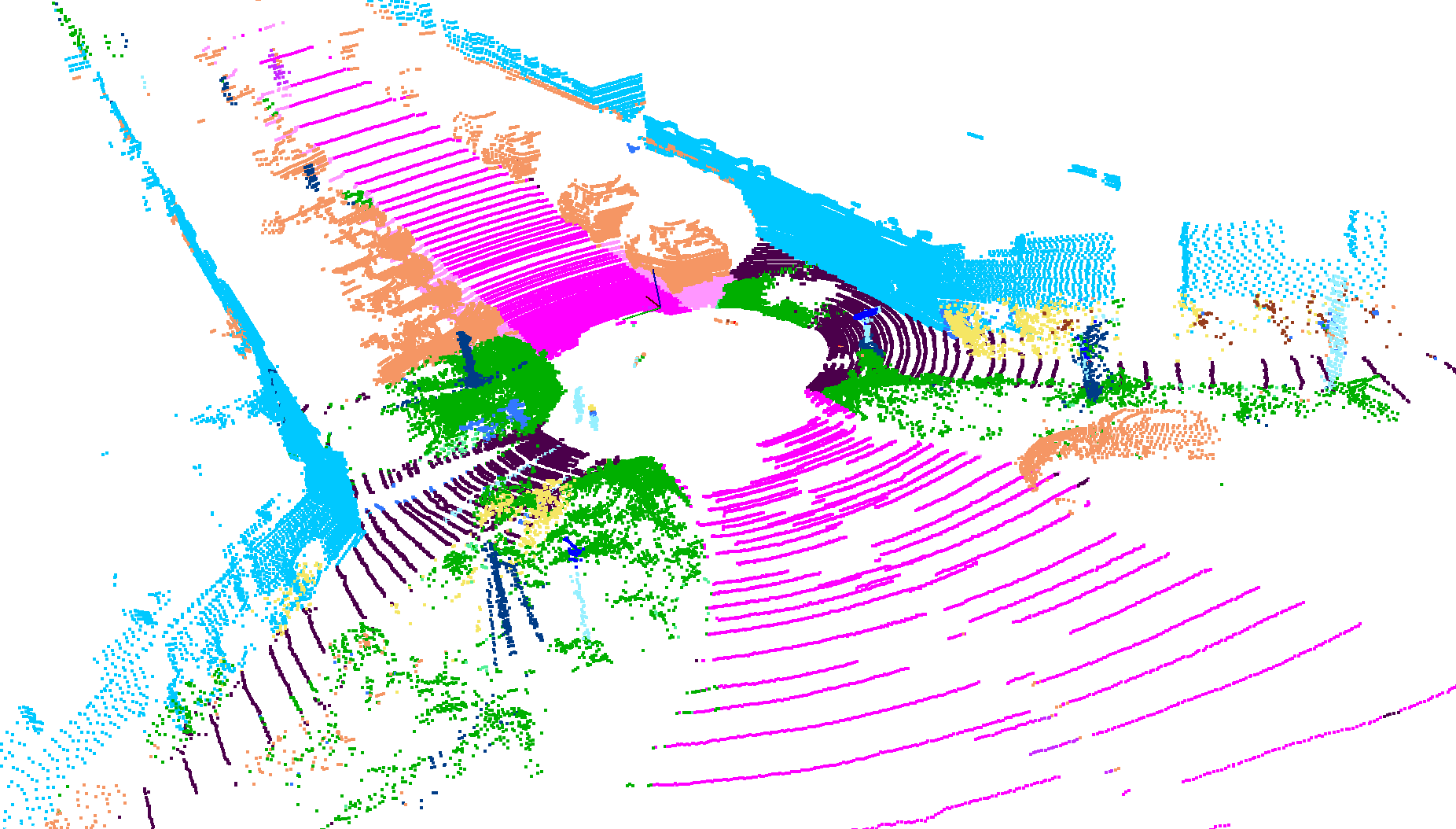}
            }
    \subfigure[Ford sequence 01]{
        \centering
        \includegraphics[width=0.45\columnwidth]{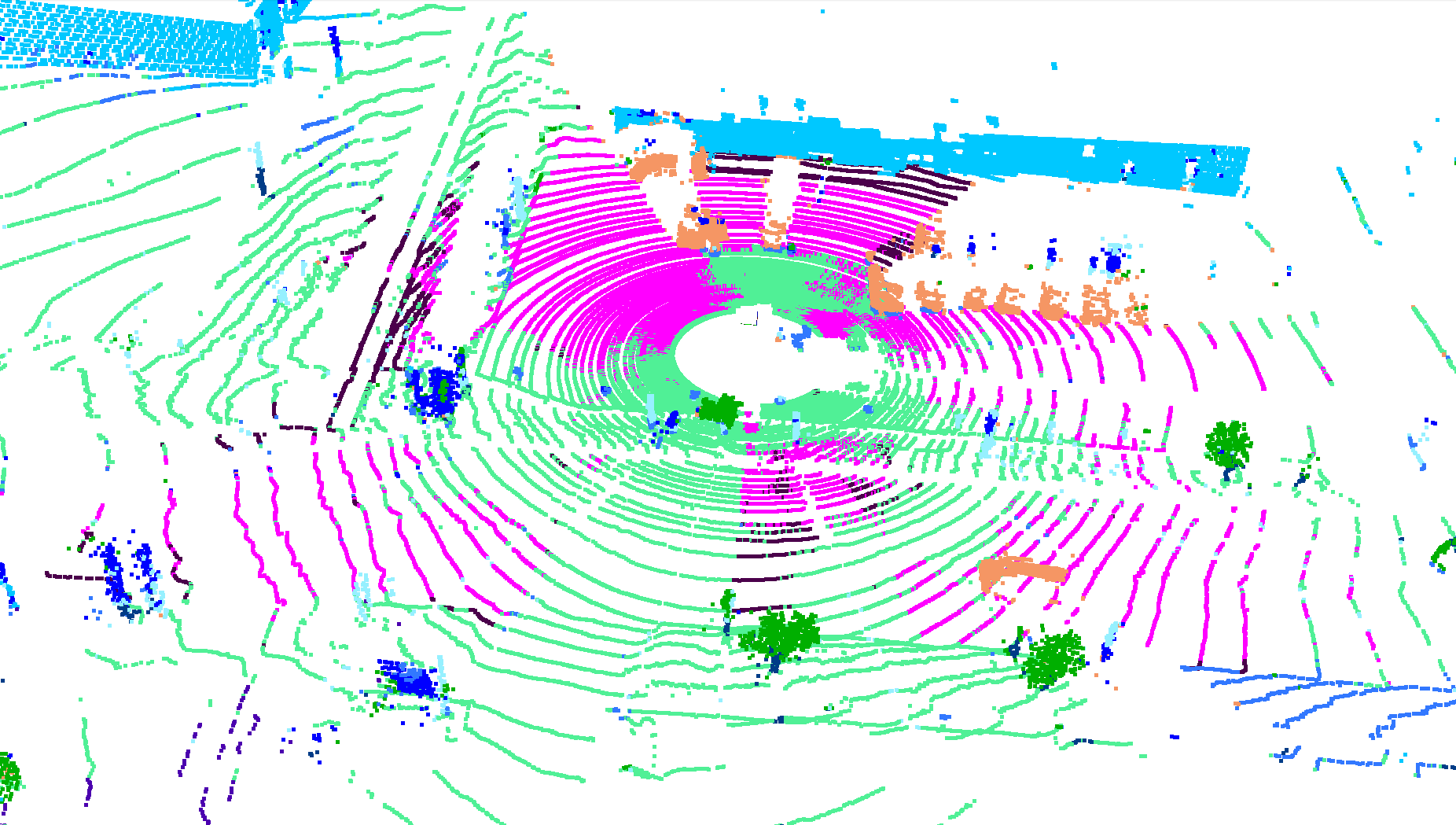}
        }
        \vspace{-10pt}
    \caption{Semantic segmentation results on KITTI and Ford.}
    \label{pic:semantic_segmentation}
 \end{figure}

Tab.~\ref{table:ford_abs} shows the ATE, and Ours-LOOP outperforms others on both sequences. What's more, among all the methods, only Ours-LOOP successfully correct the loop closures shown in Fig.~\ref{pic:ford_relative}. Note that the semantic segmentation on Ford is much worse than on KITTI, shown in Fig.~\ref{pic:semantic_segmentation}, which will inevitably affect the subsequent performance. SUMA++ and our methods use the same semantics while ours perform more consistently, benefiting from semantic graph representation. 

\section{Conclusion}

In this paper, we present a semantic-aided LiDAR SLAM system with loop closure. We exploit semantic information to improve the accuracy of point cloud registration and design a semantic-graph-based loop closure detection module to eliminate the accumulated error. The evaluation results on KITTI Odometry Benchmark show that our semantic-based ICP can effectively improve the odometry accuracy, and our semantic-based loop closure detection is crucial for constructing a globally consistent map even in large-scale scenes. Experiments on Ford Campus Dataset indicate that our method can generalize to unseen 3D scenes, achieving state-of-the-art performance.
We hope this work can draw researchers' attention in further integrating semantics with the LiDAR SLAM system. In future work, we intend to investigate more applications based on semantic graph maps.


\bibliographystyle{ieeetr}
\bibliography{main}

\begin{thebibliography}{10}

\bibitem{zhang2014loam}
J.~Zhang and S.~Singh, ``Loam: Lidar odometry and mapping in real-time.,'' in
  {\em Robotics: Science and Systems}, vol.~2, 2014.

\bibitem{shan2018lego}
T.~Shan and B.~Englot, ``Lego-loam: Lightweight and ground-optimized lidar
  odometry and mapping on variable terrain,'' in {\em 2018 IEEE/RSJ
  International Conference on Intelligent Robots and Systems (IROS)},
  pp.~4758--4765, IEEE, 2018.

\bibitem{chen2020sloam}
S.~W. Chen, G.~V. Nardari, E.~S. Lee, C.~Qu, X.~Liu, R.~A.~F. Romero, and
  V.~Kumar, ``Sloam: Semantic lidar odometry and mapping for forest
  inventory,'' {\em IEEE Robotics and Automation Letters}, vol.~5, no.~2,
  pp.~612--619, 2020.

\bibitem{7139486}
J.~{Zhang} and S.~{Singh}, ``Visual-lidar odometry and mapping: low-drift,
  robust, and fast,'' in {\em 2015 IEEE International Conference on Robotics
  and Automation (ICRA)}, pp.~2174--2181, 2015.

\bibitem{8421015}
B.~{Bescos}, J.~M. {Fácil}, J.~{Civera}, and J.~{Neira}, ``Dynaslam: Tracking,
  mapping, and inpainting in dynamic scenes,'' {\em IEEE Robotics and
  Automation Letters}, vol.~3, no.~4, pp.~4076--4083, 2018.

\bibitem{10.1007/978-3-030-01225-0_15}
K.-N. Lianos, J.~L. Sch{\"o}nberger, M.~Pollefeys, and T.~Sattler, ``Vso:
  Visual semantic odometry,'' in {\em Computer Vision -- ECCV 2018}
  (V.~Ferrari, M.~Hebert, C.~Sminchisescu, and Y.~Weiss, eds.), (Cham),
  pp.~246--263, Springer International Publishing, 2018.

\bibitem{milioto2019rangenet++}
A.~Milioto, I.~Vizzo, J.~Behley, and C.~Stachniss, ``Rangenet++: Fast and
  accurate lidar semantic segmentation,'' in {\em Proc. of the IEEE/RSJ Intl.
  Conf. on Intelligent Robots and Systems (IROS)}, 2019.

\bibitem{hu2020randla}
Q.~Hu, B.~Yang, L.~Xie, S.~Rosa, Y.~Guo, Z.~Wang, N.~Trigoni, and A.~Markham,
  ``Randla-net: Efficient semantic segmentation of large-scale point clouds,''
  in {\em Proceedings of the IEEE/CVF Conference on Computer Vision and Pattern
  Recognition}, pp.~11108--11117, 2020.

\bibitem{thomas2019KPConv}
H.~Thomas, C.~R. Qi, J.-E. Deschaud, B.~Marcotegui, F.~Goulette, and L.~J.
  Guibas, ``Kpconv: Flexible and deformable convolution for point clouds,''
  {\em Proceedings of the IEEE International Conference on Computer Vision},
  2019.

\bibitem{zhou2020cylinder3d}
H.~Zhou, X.~Zhu, X.~Song, Y.~Ma, Z.~Wang, H.~Li, and D.~Lin, ``Cylinder3d: An
  effective 3d framework for driving-scene lidar semantic segmentation,'' {\em
  arXiv preprint arXiv:2008.01550}, 2020.

\bibitem{zhang12356deep}
F.~Zhang, J.~Fang, B.~Wah, and P.~Torr, ``Deep fusionnet for point cloud
  semantic segmentation,''

\bibitem{tang2020searching}
H.~Tang, Z.~Liu, S.~Zhao, Y.~Lin, J.~Lin, H.~Wang, and S.~Han, ``Searching
  efficient 3d architectures with sparse point-voxel convolution,'' in {\em
  European Conference on Computer Vision}, 2020.

\bibitem{cortinhal2020salsanext}
T.~Cortinhal, G.~Tzelepis, and E.~E. Aksoy, ``Salsanext: Fast,
  uncertainty-aware semantic segmentation of lidar point clouds for autonomous
  driving,'' {\em arXiv preprint arXiv:2003.03653}, 2020.

\bibitem{chen2019suma++}
X.~Chen, A.~Milioto, E.~Palazzolo, P.~Gigu{\`e}re, J.~Behley, and C.~Stachniss,
  ``Suma++: Efficient lidar-based semantic slam,'' in {\em 2019 IEEE/RSJ
  International Conference on Intelligent Robots and Systems (IROS)},
  pp.~4530--4537, IEEE, 2019.

\bibitem{behley2018efficient}
J.~Behley and C.~Stachniss, ``Efficient surfel-based slam using 3d laser range
  data in urban environments.,'' in {\em Robotics: Science and Systems}, 2018.

\bibitem{kong2020semantic}
X.~Kong, X.~Yang, G.~Zhai, X.~Zhao, X.~Zeng, M.~Wang, Y.~Liu, W.~Li, and
  F.~Wen, ``Semantic graph based place recognition for 3d point clouds,'' {\em
  arXiv preprint arXiv:2008.11459}, 2020.

\bibitem{chen2020overlapnet}
X.~Chen, T.~L{\"a}be, A.~Milioto, T.~R{\"o}hling, O.~Vysotska, A.~Haag,
  J.~Behley, C.~Stachniss, and F.~Fraunhofer, ``Overlapnet: Loop closing for
  lidar-based slam,'' in {\em Proc. of Robotics: Science and Systems (RSS)},
  2020.

\bibitem{wang2020floam}
W.~Han, ``Fast loam.'' Website, 2020.
\newblock \url{https://github.com/wh200720041/floam}.

\bibitem{zhang2017low}
J.~Zhang and S.~Singh, ``Low-drift and real-time lidar odometry and mapping,''
  {\em Autonomous Robots}, vol.~41, no.~2, pp.~401--416, 2017.

\bibitem{geiger2013vision}
A.~Geiger, P.~Lenz, C.~Stiller, and R.~Urtasun, ``Vision meets robotics: The
  kitti dataset,'' {\em The International Journal of Robotics Research},
  vol.~32, no.~11, pp.~1231--1237, 2013.

\bibitem{deschaud2018imls}
J.-E. Deschaud, ``Imls-slam: scan-to-model matching based on 3d data,'' in {\em
  2018 IEEE International Conference on Robotics and Automation (ICRA)},
  pp.~2480--2485, IEEE, 2018.

\bibitem{8387438}
A.~{Zaganidis}, L.~{Sun}, T.~{Duckett}, and G.~{Cielniak}, ``Integrating deep
  semantic segmentation into 3-d point cloud registration,'' {\em IEEE Robotics
  and Automation Letters}, vol.~3, no.~4, pp.~2942--2949, 2018.

\bibitem{biber2003normal}
P.~Biber and W.~Stra{\ss}er, ``The normal distributions transform: A new
  approach to laser scan matching,'' in {\em Proceedings 2003 IEEE/RSJ
  International Conference on Intelligent Robots and Systems (IROS 2003)(Cat.
  No. 03CH37453)}, vol.~3, pp.~2743--2748, IEEE, 2003.

\bibitem{segal2009generalized}
A.~Segal, D.~Haehnel, and S.~Thrun, ``Generalized-icp.,'' in {\em Robotics:
  science and systems}, vol.~2, p.~435, Seattle, WA, 2009.

\bibitem{johnson1999using}
A.~E. Johnson and M.~Hebert, ``Using spin images for efficient object
  recognition in cluttered 3d scenes,'' {\em IEEE Transactions on pattern
  analysis and machine intelligence}, vol.~21, no.~5, pp.~433--449, 1999.

\bibitem{salti2014shot}
S.~Salti, F.~Tombari, and L.~Di~Stefano, ``Shot: Unique signatures of
  histograms for surface and texture description,'' {\em Computer Vision and
  Image Understanding}, vol.~125, pp.~251--264, 2014.

\bibitem{tombari2011combined}
F.~Tombari, S.~Salti, and L.~Di~Stefano, ``A combined texture-shape descriptor
  for enhanced 3d feature matching,'' in {\em 2011 18th IEEE international
  conference on image processing}, pp.~809--812, IEEE, 2011.

\bibitem{guo2019local}
J.~Guo, P.~V. Borges, C.~Park, and A.~Gawel, ``Local descriptor for robust
  place recognition using lidar intensity,'' {\em IEEE Robotics and Automation
  Letters}, vol.~4, no.~2, pp.~1470--1477, 2019.

\bibitem{rusu2009fast}
R.~B. Rusu, N.~Blodow, and M.~Beetz, ``Fast point feature histograms (fpfh) for
  3d registration,'' in {\em 2009 IEEE international conference on robotics and
  automation}, pp.~3212--3217, IEEE, 2009.

\bibitem{he2016m2dp}
L.~He, X.~Wang, and H.~Zhang, ``M2dp: A novel 3d point cloud descriptor and its
  application in loop closure detection,'' in {\em 2016 IEEE/RSJ International
  Conference on Intelligent Robots and Systems (IROS)}, pp.~231--237, IEEE,
  2016.

\bibitem{Kim2018ScanCE}
G.~Kim and A.~Kim, ``Scan context: Egocentric spatial descriptor for place
  recognition within 3d point cloud map,'' {\em 2018 IEEE/RSJ International
  Conference on Intelligent Robots and Systems (IROS)}, pp.~4802--4809, 2018.

\bibitem{8500682}
H.~{Yin}, L.~{Tang}, X.~{Ding}, Y.~{Wang}, and R.~{Xiong}, ``Locnet: Global
  localization in 3d point clouds for mobile vehicles,'' in {\em 2018 IEEE
  Intelligent Vehicles Symposium (IV)}, pp.~728--733, 2018.

\bibitem{9196764}
H.~{Wang}, C.~{Wang}, and L.~{Xie}, ``Intensity scan context: Coding intensity
  and geometry relations for loop closure detection,'' in {\em 2020 IEEE
  International Conference on Robotics and Automation (ICRA)}, pp.~2095--2101,
  2020.

\bibitem{angelina2018pointnetvlad}
M.~Angelina~Uy and G.~Hee~Lee, ``Pointnetvlad: Deep point cloud based retrieval
  for large-scale place recognition,'' in {\em Proceedings of the IEEE
  Conference on Computer Vision and Pattern Recognition}, pp.~4470--4479, 2018.

\bibitem{liu2019seqlpd}
Z.~Liu, C.~Suo, S.~Zhou, F.~Xu, H.~Wei, W.~Chen, H.~Wang, X.~Liang, and Y.-H.
  Liu, ``Seqlpd: Sequence matching enhanced loop-closure detection based on
  large-scale point cloud description for self-driving vehicles,'' in {\em 2019
  IEEE/RSJ International Conference on Intelligent Robots and Systems (IROS)},
  pp.~1218--1223, IEEE, 2019.

\bibitem{liu2019lpd}
Z.~Liu, S.~Zhou, C.~Suo, P.~Yin, W.~Chen, H.~Wang, H.~Li, and Y.-H. Liu,
  ``Lpd-net: 3d point cloud learning for large-scale place recognition and
  environment analysis,'' in {\em Proceedings of the IEEE International
  Conference on Computer Vision}, pp.~2831--2840, 2019.

\bibitem{qi2017pointnet}
C.~R. Qi, H.~Su, K.~Mo, and L.~J. Guibas, ``Pointnet: Deep learning on point
  sets for 3d classification and segmentation,'' in {\em Proceedings of the
  IEEE conference on computer vision and pattern recognition}, pp.~652--660,
  2017.

\bibitem{qi2017pointnet++}
C.~R. Qi, L.~Yi, H.~Su, and L.~J. Guibas, ``Pointnet++: Deep hierarchical
  feature learning on point sets in a metric space,'' in {\em Advances in
  neural information processing systems}, pp.~5099--5108, 2017.

\bibitem{arandjelovic2016netvlad}
R.~Arandjelovic, P.~Gronat, A.~Torii, T.~Pajdla, and J.~Sivic, ``Netvlad: Cnn
  architecture for weakly supervised place recognition,'' in {\em Proceedings
  of the IEEE conference on computer vision and pattern recognition},
  pp.~5297--5307, 2016.

\bibitem{dube2017segmatch}
R.~Dub{\'e}, D.~Dugas, E.~Stumm, J.~Nieto, R.~Siegwart, and C.~Cadena,
  ``Segmatch: Segment based place recognition in 3d point clouds,'' in {\em
  2017 IEEE International Conference on Robotics and Automation (ICRA)},
  pp.~5266--5272, IEEE, 2017.

\bibitem{segmap2018}
R.~Dub{\'e}, A.~Cramariuc, D.~Dugas, J.~Nieto, R.~Siegwart, and C.~Cadena,
  ``{SegMap}: 3d segment mapping using data-driven descriptors,'' in {\em
  Robotics: Science and Systems (RSS)}, 2018.

\bibitem{dube2019segmap}
R.~Dub{\'e}, A.~Cramariuc, D.~Dugas, H.~Sommer, M.~Dymczyk, J.~Nieto,
  R.~Siegwart, and C.~Cadena, ``Segmap: Segment-based mapping and localization
  using data-driven descriptors,'' {\em The International Journal of Robotics
  Research}, p.~0278364919863090, 2019.

\bibitem{inproceedings}
R.~Kümmerle, G.~Grisetti, H.~Strasdat, K.~Konolige, and W.~Burgard, ``G2o: A
  general framework for graph optimization,'' pp.~3607 -- 3613, 06 2011.

\bibitem{pandey2011ford}
G.~Pandey, J.~R. McBride, and R.~M. Eustice, ``Ford campus vision and lidar
  data set,'' {\em The International Journal of Robotics Research}, vol.~30,
  no.~13, pp.~1543--1552, 2011.

\bibitem{paszke2019pytorch}
A.~Paszke, S.~Gross, F.~Massa, A.~Lerer, J.~Bradbury, G.~Chanan, T.~Killeen,
  Z.~Lin, N.~Gimelshein, L.~Antiga, {\em et~al.}, ``Pytorch: An imperative
  style, high-performance deep learning library,'' in {\em Advances in Neural
  Information Processing Systems}, pp.~8024--8035, 2019.

\bibitem{behley2019semantickitti}
J.~Behley, M.~Garbade, A.~Milioto, J.~Quenzel, S.~Behnke, C.~Stachniss, and
  J.~Gall, ``Semantickitti: A dataset for semantic scene understanding of lidar
  sequences,'' in {\em Proceedings of the IEEE International Conference on
  Computer Vision}, pp.~9297--9307, 2019.

\end{thebibliography}


\end{document}